\documentclass[letterpaper, 10pt, conference]{ieeeconf}
\IEEEoverridecommandlockouts 
\usepackage[utf8]{inputenc}
\usepackage{hyperref}
\usepackage{amsmath}
\usepackage{amssymb}
\usepackage{graphicx}
\usepackage{dsfont}
\usepackage{todonotes}
\usepackage[T1]{fontenc}
\usepackage{textcomp}
\usepackage{subfig}

\newcommand{\etal}{\textit{et al}. }

\title{\LARGE \bf Any Way You Look At It: Semantic\\Crossview Localization and Mapping with LiDAR}
\author{Ian D. Miller$^{1}$, Anthony Cowley$^{1}$, Ravi Konkimalla$^{1}$, Shreyas S. Shivakumar$^{1}$,\\Ty Nguyen$^{1}$, Trey Smith$^{2}$, Camillo Jose Taylor$^{1}$, and Vijay Kumar$^{1}$
\thanks{We gratefully acknowledge the support from ARL Grant DCIST CRA W911NF-17-2-0181, NSF Grant CNS-1521617,
ARO Grant W911NF-13-1-0350, ONR Grants N00014-20-1-2822 and
ONR grant N00014-20-S-B001,
Qualcomm Research,
and C-BRIC, a Semiconductor Research Corporation Joint University Microelectronics Program  program cosponsored by DARPA.
Ian Miller acknowledges support from a NASA Space Technology Research Fellowship.
}
\thanks{$^{1}$ Ian D. Miller, Anthony Cowley, Ravi Konkimalla, Shreyas S. Shivakumar, Ty Nguyen, Camillo Jose Taylor, and Vijay Kumar are with the
GRASP Lab, University of Pennsylvania, Philadelphia, PA 19104.  Corresponding author: \tt\footnotesize iandm@seas.upenn.edu}
\thanks{$^{2}$ Trey Smith is with the NASA Ames Intelligent Robotics Group, Moffett Field, CA 94035
}
}

\newcommand\copyrighttext{%
  \footnotesize \textcopyright 2021 IEEE. Personal use of this material is permitted.
  Permission from IEEE must be obtained for all other uses, in any current or future
  media, including reprinting/republishing this material for advertising or promotional
  purposes, creating new collective works, for resale or redistribution to servers or
  lists, or reuse of any copyrighted component of this work in other works.
  DOI: \href{https://ieeexplore.ieee.org/document/9361130}{10.1109/LRA.2021.3061332}}
\newcommand\copyrightnotice{%
\begin{tikzpicture}[remember picture,overlay]
\node[anchor=south,yshift=10pt] at (current page.south) {\fbox{\parbox{\dimexpr\textwidth-\fboxsep-\fboxrule\relax}{\copyrighttext}}};
\end{tikzpicture}%
}

\begin{document}

\maketitle
\copyrightnotice

\begin{abstract}
    Currently, GPS is by far the most popular global localization method.  However, it is not always reliable or accurate in all environments.  SLAM methods enable local state estimation but provide no means of registering the local map to a global one, which can be important for inter-robot collaboration or human interaction.  In this work, we present a real-time method for utilizing semantics to globally localize a robot using only egocentric 3D semantically labelled LiDAR and IMU as well as top-down RGB images obtained from satellites or aerial robots.  Additionally, as it runs, our method builds a globally registered, semantic map of the environment.  We validate our method on KITTI as well as our own challenging datasets, and show better than 10 meter accuracy, a high degree of robustness, and the ability to estimate the scale of a top-down map on the fly if it is initially unknown.
\end{abstract}

\section{Introduction}

Localization is a fundamental problem in mobile robotics. From self-driving cars \cite{ma_icra_2017} to exploratory Micro Aerial Vehicles (MAVs) \cite{tung_aero_2020}, robots need to know where they are.  This problem is particularly challenging for multi-robot systems, a setting in which effective collaboration not only typically assumes a shared understanding of the global map \cite{gawel_isssrr_2017}, but also presents a significant opportunity given the unique information each agent possesses.  The most common solutions to this problem are GPS \cite{peterson_sensors_2018} or agreed-upon starting configurations \cite{michael_jfr_2012}.  However, GPS is often unreliable or unavailable, and approaches that unify local maps based on the starting configuration are susceptible to drift over time and can be sensitive to small errors in the initial alignment.  Other options include boostrapping robot-robot detections~\cite{liang_tro_2018} with visual inertial odometry to localize robots within a common moving frame~\cite{franchi2013mutual}, but these approaches fail if robots are not in each other's field of view or are too far away.

The relative localization problem can be solved neatly if robots are capable of localizing themselves in each other's maps.  This problem amounts to the loop-closure problem for heterogeneous robots with similar sensing modalities \cite{howard_ijrr_2006} or even different sensors \cite{wang_sensors_2019}, provided the robots move in a similar space.  However, if the robots' trajectories never overlap, loop closures cannot help.  Therefore, we propose registering robots' local measurements against larger global maps from satellite or overhead MAV imagery. This approach is more challenging than conventional known-map localization methods \cite{dellaert_icra_1999, majdik_iros_2013} because the overhead views are very different from the egocentric data available to most robots \cite{gao_prl_2019}. Therefore, it is ineffective to directly apply feature-based registration approaches to localize robots in these aerial views.  Additionally, satellite views may have been captured at different times, meaning registration must cope with seasonal variations.  We must therefore efficiently compare the satellite and egocentric perspectives.

\begin{figure}
    \centering
    \includegraphics[width=0.8\linewidth]{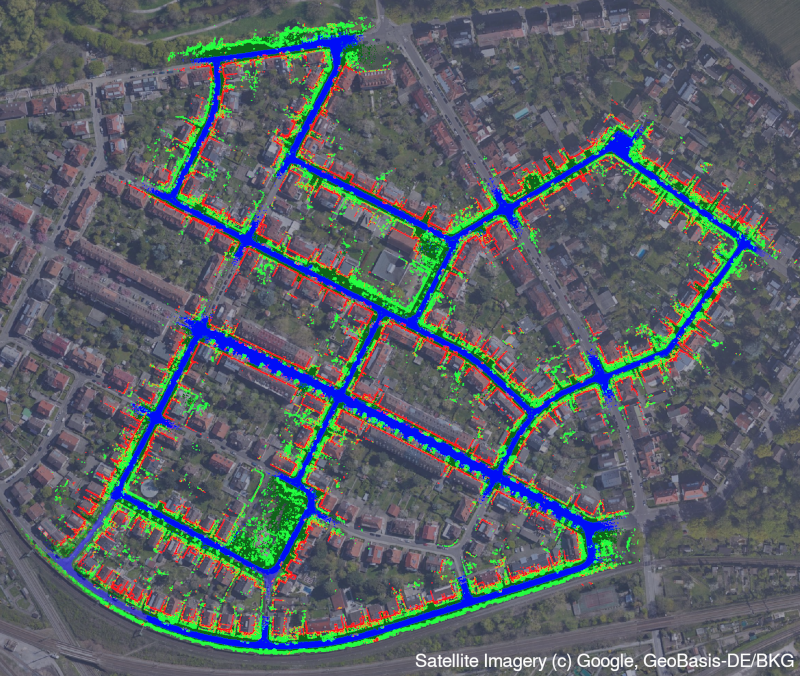}
    \caption{Georeferenced, semantic pointcloud of a KITTI dataset labelled and mapped in realtime overlaid on satellite imagery.}
    \label{fig:kitti0_top_down}
    \vspace{-5mm}
\end{figure}

In recent years, image semantic segmentation using artificial neural networks (ANNs) has become a mature technology \cite{wang_corr_2019}.  Map semantics are ideal for cross-view registration since they are invariant to perspective and seasonal changes given sufficiently varied training data. In addition, coarse semantic aerial maps are comparatively easy to generate from overhead imagery, either manually or using additional ANNs \cite{wu_access_2019}.  At the same time, LiDARs have increased in beam density and decreased in price and weight, so placing them on small mobile robots is now practical.  The fusion of these two sensing technologies leads to dense semantic point clouds, which we suggest are an ideal modality for the cross-view localization problem.

Recent works have used image semantics for cross view localization \cite{castaldo_iccv_2015, gawel_ral_2018}, but typically do not or minimally utilize depth information or make strong structural assumptions about the environment.  Geometric methods such as \cite{gawel_isssrr_2017} fail in environments without sufficient geometric structure.  We instead, introduce an approach which combines these two sources of information to enable a more robust semantic cross-localization system. In addition, our system is entirely environmentally agnostic, assuming only some predetermined set of classes.

Our contributions in this work are as follows:
\begin{enumerate}
    \item We present a real-time cross-view localization and mapping framework using semantic point clouds for localization to an overhead map. Our method is also capable of estimating the scale of the aerial map if it is unknown but bounded.
    \item We validate our proposed method using both SemanticKITTI \cite{behley_iccv_2019} as well as our own datasets in a variety of locations including rural and urban environments.  Validation is additionally performed using inferred, noisy, real segmentation data.
    \item We release our datasets and localization code in order to further work in the field by the broader research community\footnote{\tt\footnotesize https://github.com/iandouglas96/cross\_view\_slam}.
\end{enumerate}

\section{Related Work}

\subsection{Image Matching}

Cross-view localization has been formulated as an image matching problem
\cite{gao_prl_2019}.  Given a database of images that represent the global map and a query image representing the local observation, the problem can be formulated as finding some descriptor such that images of the same location from a diversity of views are close in some latent space.  Early works such as \cite{chen_cvpr_2011}, \cite{li_eccv_2010} use local feature descriptors for this purpose. Majdik \etal \cite{majdik_iros_2013} use such features with simulated images from Google Street View to match against images from a quadrotor flying through an urban environment.  However, these methods fail with more extreme viewpoint variation between the query and database images.

This limitation has led to recent interest in siamese networks \cite{kim_icra_2017,  tian2020cross}.  These networks perform cross correlation between the latent representations of two viewpoints given a weight-sharing branch for each viewpoint in order to determine how close the two viewpoints are. The whole network is trained on known pairs of images with a loss encouraging a common representation in latent space.  While achieving impressive accuracy, these networks are relatively slow in processing an image pair, and therefore cannot be run in real time in robotics applications where many possible states must be queried rapidly simultaneously.

\subsection{Vision-based State Estimation}

Image matching algorithms provide a means of comparing an image against a provided database of existing images, but they do not explicitly seek to estimate the position of the robot or sensor.  
Localization requires either an image database with pose labels or an entire aerial map as in \cite{leung_icra_2008}.  In this work, the authors represent the global map as a series of edges in the aerial frame, which are then matched against edges in ground-level imagery in a particle filter framework. However, by reducing the global map to a set of edges, the method discards a large quantity of potentially useful information. Other works such as \cite{senlet_iccv_2011} use stereo imagery to generate an RGBD image, which they then render from a aerial perspective and compare with the known map using chamfer matching. This approach nonetheless fails to address the photographic changes caused by factors like seasonal changes or dynamic objects in the environment such as people or cars. 

In order to better cope with seasonal changes and more extreme viewpoint variation, recent work has increasingly emphasized the use of semantics for localization.  Castaldo \etal \cite{castaldo_iccv_2015} segment ground-based images and use the homography with the ground plane to project these semantics into a top-down view.  They then develop a semantic descriptor for the segmented image and compare with the known map to generate a heatmap over the set of camera locations.  However, they do not utilize temporal or depth information, causing slow convergence in a large-scale localization system.  Their projection method also fails for cases where the homography assumption does not hold, such as the case where the region in front of the robot is not close to flat.
In a similar work, Gawel \etal \cite{gawel_ral_2018} generate a semantic graph representation for both aerial and ground-view images which they then build descriptors on to match in a variety of synthetic datasets.  However, the method minimally exploits geometric information by reducing the semantic segmentation to the centroids of class blobs, and therefore fails for environments with few classes and minimal variety. 

\subsection{LiDAR-based State Estimation}

It is intuitively clear that projecting data between frames becomes easier in the presence of depth information. Wolcott \etal \cite{wolcott_iros_2014} localize a camera inside a known high-definition pointcloud by rendering the cloud from various perspectives and maximizing the mutual information with a given camera image over the state space.  Gawel \etal \cite{gawel_isssrr_2017} cross-localize ground and aerial robots by building point cloud maps from both perspectives and matching them using geometric descriptors. Barsan \etal \cite{barsan_pmlr_2018} achieve centimeter-level localization by using a siamese network with top-down views of LiDAR intensity scans.  While accurate and powerful, these methods require a preexisting point cloud map of the relevant environment, while we require only a single aerial image.

As with vision-based methods, state-of-the-art LiDAR methods exploit the semantic structure of environments.  In an early work, Matei \etal \cite{matei_wacv_2013} fit closed building models to a point cloud gathered from an aerial vehicle which are used to create ground-level predictions to compare to the ground imagery.  Senlet \etal \cite{senlet_icra_2014} match local top-down views to large satellite maps by segmenting buildings with an ANN and comparing hashed descriptors that capture the local building layout.
Tian \etal \cite{tian_cvpr_2017} use a similar hybrid method which also segments buildings, but they utilize a siamese network to generate building descriptors. 
Similar to ours, Yan \etal~\cite{yan_ecmr_2019} perform 3D LiDAR-based localization using a simple top-down view map of the environment from OpenStreetMap. The authors build a 4-bit descriptor for matching which incorporates information about intersections and building gaps.  While this approach is highly efficient and can run in real-time, it must throw out large quantities of data in order to compress a 3D pointcloud to 4 bits. 
These methods all assume strong priors on the structure of the localization environment and fail in general (i.e. non-urban) environments.


More general methods such as \cite{stenborg_icra_2018, liu_icra_2019} utilize a 3D labeled point map of the environment. In~\cite{stenborg_icra_2018}, registration is performed by comparing the projected pointcloud to egocentric images from the robot while in~\cite{liu_icra_2019}, the authors use random walk descriptors for the matching.
Compared to these works, our framework requires only a single satellite or aerial view of the environment, making it suitable to more applications. 



 
\section{Method}

\begin{figure}
    \vspace{2mm}
    \centering
    \includegraphics[width=0.45\textwidth]{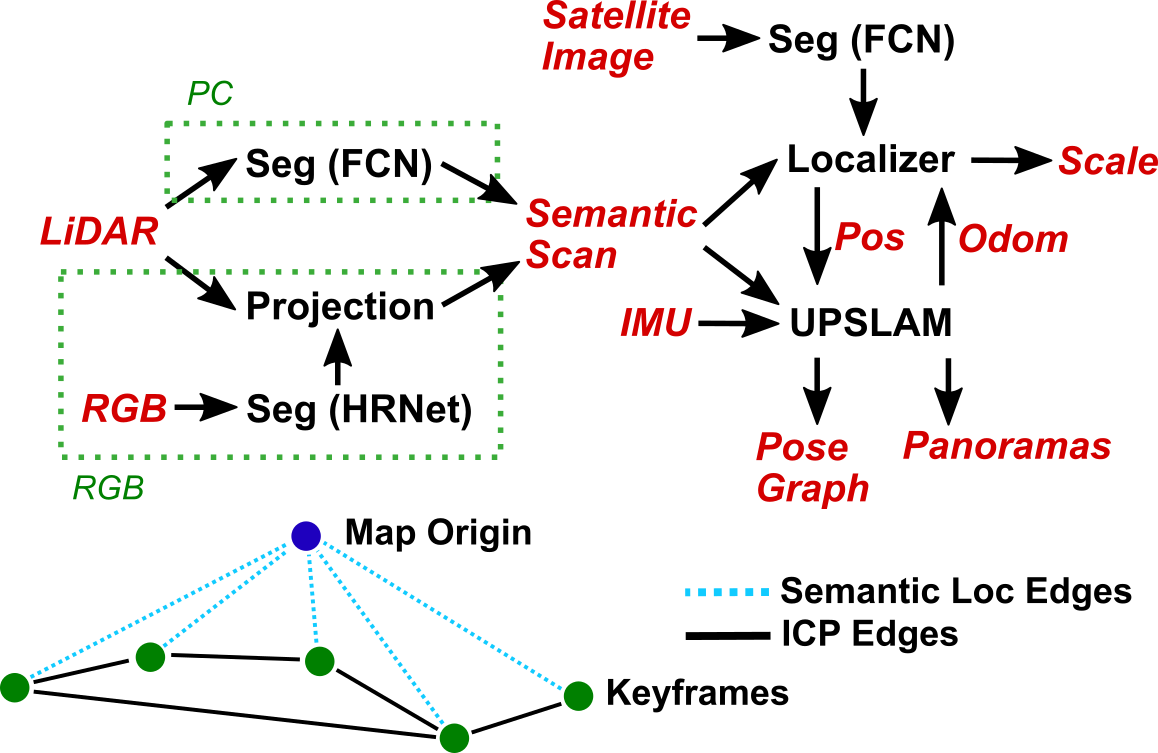}
    \caption{Block diagram of the overall system (top) showing each optional data pipeline in a dotted green box.  Sample pose graph (bottom) showing the supplemented UPSLAM map.}
    \label{fig:block_diag}
    \vspace{-5mm}
\end{figure}

Our method consists of two primary components: an ICP-based LiDAR SLAM system---Union of Panoramas SLAM \cite{cowley2021upslam} (UPSLAM)---, and a particle-filter based semantic localizer as shown in Fig.~\ref{fig:block_diag}.  Here we detail the purpose and implementation of both of these systems.

\subsection{Localization}
Inspired by the success of Monte-Carlo methods for robot localization \cite{yan_ecmr_2019, leung_icra_2008, dellaert_icra_1999}, we adopt a particle filter for the backbone of our localization pipeline.  Particle filters are particularly adept at handling multimodal state distributions, which appear frequently in robot localization problems.  This comes, however, at a computational cost, which we mitigate by employing customized optimization strategies.

\subsubsection{Problem Formulation}
For 2D map localization our system state $x$ consists of the tuple $(p \in SE(2), s \in [s_{\text{min}}, s_{\text{max}}])$, where $p$ is the robot's orientation and position (in meters) on the map and $s$ (in px/m) is the map scale with some prior on its bounds.  We additionally have the input $u \in SE(3)$, the rigid-body transformation of the robot from the last frame (in meters) from UPSLAM or any other odometry source.  We are then left with the problem of defining the motion model $P(x_t \mid x_{t-1}, u_{t-1})$ for every timestep $t$.  In addition, at every $t$ we obtain the semantic scan $z_t$ from our LiDAR and cameras.  In order to define our particle filter, we must also define the measurement model $P(z_t \mid x_t)$.

\subsubsection{Motion Model}
We do not assume access to the control inputs to the robot and instead rely on frame-to-frame motion estimates from UPSLAM.  Because UPSLAM operates in 3D we first project the frame-to-frame motion onto the local x-y ground plane, an operation we denote as $\text{proj}(u) \in SE(2)$.  In addition, we use only the initial ICP solution from UPSLAM for our motion estimation, ignoring the optimized pose from loop closures in order to avoid discontinuities in odometry.  Finally, we assume normally distributed noise with constant covariance for each estimate, with scale noise existing in log-space.  In summary,
\begin{equation}
    \begin{split}
        P(x_t \mid x_{t-1}, u_{t-1}) = (&[\text{proj}(u_{t-1}) + \mathcal{N}(0, \Sigma_p)] * p_{t-1}, \\
        &\mathcal{N}(1, \Sigma_s) * s_{t-1})
    \end{split}
\end{equation}
We additionally scale $\Sigma_s$ with the inverse of the distance from the starting position in order to encourage natural convergence.  Once the scale variance in the state drops below a threshold we fix $s$.

\subsubsection{Measurement Model}
Our measurement, $z$, is a semantic pointcloud, which we can represent as a list of point locations in the robot frame with associated labels: $z = \{(p_1, l_1), (p_2, l_2), ... (p_n, l_n)\}$.  We can then project the points onto the ground plane.  Furthermore, for any given particle state, we can query the top-down aerial map $L$ for the expected class of any point in the robot frame.  A simple way to compute a cost for a particular particle with pose $d$ would then be
\begin{equation}
    C' = \sum_{i \in [1, n]} \mathbf{1}(L(d * p_i) \neq l_i).
\end{equation}

In order to improve convergence by enlarging the local minimum, we pick a softer cost-function.  Instead of assessing cost in a binary fashion, we penalize the (thresholded) distance of a point of particular class from the closest point of the same class on the aerial map.  The cost function then becomes
\begin{equation}
    C = \sum_{i \in [1, n]} \min_{\{p \mid L(p) = l_i\}}(||p - p_i||)
    \label{eq:theoretical_cost}
\end{equation}

Finally, we compute an ad-hoc probability by inverting $C$ and normalizing.  We additionally introduce a per-class weighting factor $\alpha_l$.  That is,
\begin{equation}
    P(z_t \mid x_t) \approx \frac{n}{\sum_{i \in [1, n]} \min_{\{p \mid L(p) = l_i\}} (\alpha_{l_i} ||p - p_i||)} + \gamma
\end{equation}
where $\gamma$ is a regularization constant introduced to slow convergence.  Probabilities for all particles are finally normalized such that their sum over all particles is 1.  We note that this is an ad-hoc measure, and in practice constants are tuned through experimentation.  However, this is not uncommon for Monte-Carlo based localization methods such as \cite{yan_ecmr_2019}.

\subsubsection{Performance Optimization}
\begin{figure}
    \centering
    \vspace{2mm}
    \includegraphics[width=0.85\linewidth]{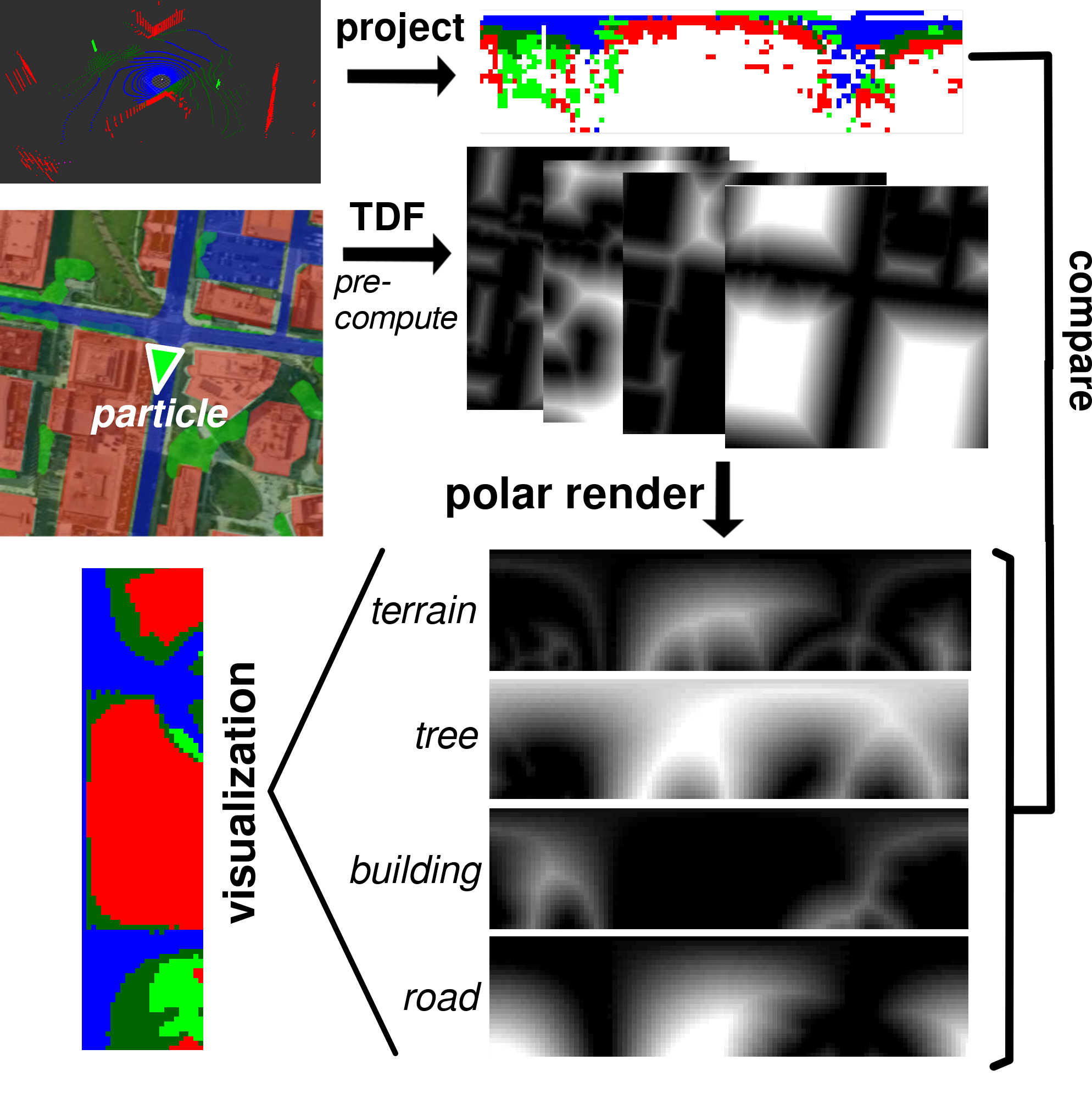}
    \caption{Processing pipeline for particle weight computation.}
    \label{fig:weighting_pipe}
    \vspace{-5mm}
\end{figure}

Naively implemented, Eq.~\ref{eq:theoretical_cost} is very expensive to compute, as it amounts to summing the results of a series of (non-convex) minimizations. We optimize this by precomputing a class-wise truncated distance field (TDF) for the aerial semantic map.  The computation is implemented very naively and takes about a minute, but only ever has to be done once per map.  This map encodes the thresholded distance from every point to the nearest point of that class, turning the minimization of Eq.~\ref{eq:theoretical_cost} into a simple fixed-time lookup.  Additionally, instead of summing over all points, we first discretize the semantic LiDAR scan into polar segments and count the number of points of each class that lie in each segment. The local class-wise truncated distance fields are rendered in the same manner. We can then approximate Eq.~\ref{eq:theoretical_cost} by summing the element-wise products of these two images resulting in one inner product operation per class, a computation that can be performed efficiently.  We found it to take approximately $500\mu\text{s}$ per particle for $100 \times 25$ pixel polar maps, with almost all of this time being used to lookup the polar TDFs.  This process is shown in Fig.~\ref{fig:weighting_pipe}.

In addition, rotating the map in the polar representation amounts to an index shift and is very fast, a fact which we exploit during initialization.  We randomly sample points on the map roads, since we have strong prior knowledge that we are beginning on a road.  For each point, we initialize $k_s$ particles of uniformly varying scale between $s_{\text{min}}$ and $s_{\text{max}}$ (1 and 10 px/m for our experiments).  For each particle, we uniformly sample $k_\theta$ possible orientations, which we can do very efficiently with index shifts.  We pick the best orientation and use this as our initial particle.

To further accelerate our algorithm, we parallelize the cost computation for each particle across the CPU cores.  We also adaptively select the number of particles based on the sum of the areas of the covariance ellipses of a Gaussian Mixture Model (GMM) fit to the particle distribution.  We expect large performance gains could be had by moving the particle weighting computation onto the GPU, but leave this as future work.

\subsection{Semantic Segmentation}  
\subsubsection{Satellite Segmentation}
We train two slightly modified versions of a Fully Convolutional Network (FCN) \cite{long2015fully} with a ResNet-34 \cite{he2016deep} backbone pretrained on ImageNet to segment the satellite images.  Images are taken directly from Google Earth.  We use 4 classes: \texttt{road}, \texttt{terrain}, \texttt{vegetation}, and \texttt{building}.
To analyze the satellite imagery $256 \times 256$ px RGB images are passed as inputs to the image segmentation network. The network was trained on the three manually labelled satellite images from the datasets in Table \ref{tab:dataset_summary}. The images are randomly scaled, rotated, cropped, and flipped to generate more training samples. The random scaling of the satellite images also allows the model to generalize better on images collected from multiple altitudes. Example output and training data of our network can be seen in Fig.~\ref{fig:labelled_maps}.  While it may seem surprising that our model performs well despite training on only 3 images, each image contains many object instances, and furthermore that a training image is from the same city as the test images, requiring minimal generalization.

\subsubsection{Scan Segmentation}
We use two different pipelines shown in green boxes in Fig.~\ref{fig:block_diag} for generating the semantic pointclouds depending on the dataset. By testing using these differing segmentation methods, we show that our method can generalize well between different sensing systems, provided that the ultimate output is a point-wise labelled pointcloud.

\textbf{PC:} For the KITTI datasets, we use the same FCN structure (also ImageNet pretrained) as for satellite segmentation, but operate on LiDAR scans represented as a 2D Polar Grid Map of size (64, 2048) with X,Y,Z, and Depth channels (we do not use intensity). We train on SemanticKITTI and use $\{10\}$ and $\{00, 02, 09\}$ as the validation and test splits, with the remaining data in the training split.  We also add additional \texttt{vehicle} and \texttt{other} classes to the set of classes used for satellite segmentation, with \texttt{vehicle} being converted to \texttt{road} as part of the top-down projection process.

\textbf{RGB:} For our own Morgantown and UCity datasets, we use a different LiDAR sensor from KITTI (an Ouster OS-1) and therefore cannot use SemanticKITTI for training data.  We instead segment RGB images calibrated to the LiDAR using HRNets \cite{wang_corr_2019} trained on Cityscapes \cite{cordts_cvpr_2016}.  Using our computed extrinsics we can then project the LiDAR pointcloud onto the camera frames and assign the appropriate class to each point from the RGB segmentation.

\subsection{Mapping}

\begin{figure}
    \vspace{2mm}
    \centering
    \includegraphics[width=0.8\linewidth]{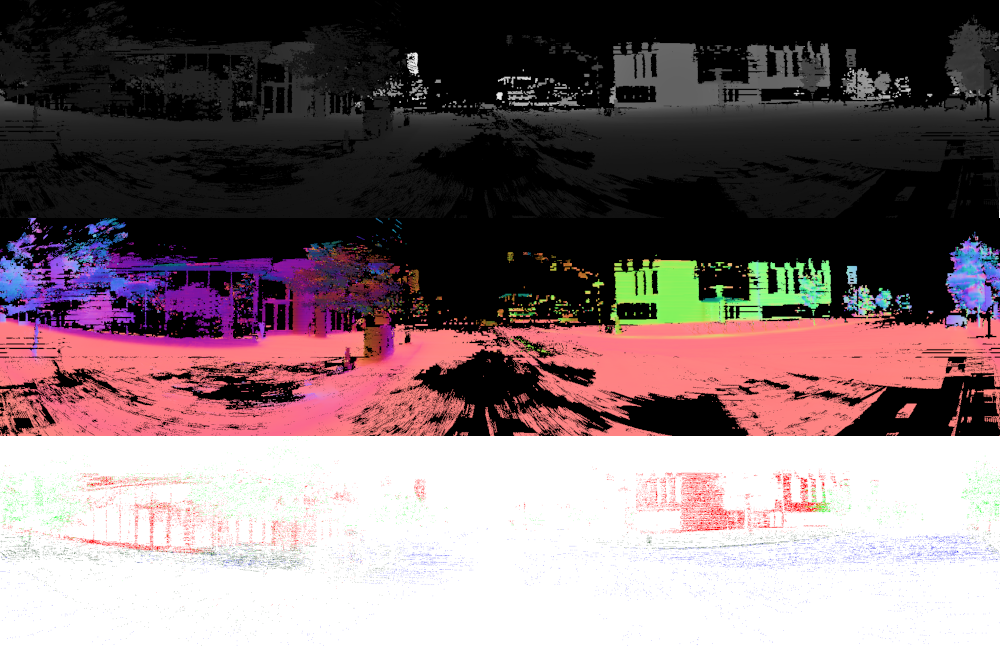}
    \caption{Depth, Normal, and Semantic panoramas integrated by UPSLAM for a portion of the UCity dataset}
    \label{fig:panos}
    \vspace{-5mm}
\end{figure}

We use the UPSLAM mapper described in \cite{cowley2021upslam} but provide a brief overview of our relevant changes here.
UPSLAM estimates the motion of the LIDAR sensor using Iterative Closest Point (ICP) and represents the overall geometry in terms of a collection of panoramic depth maps attached to keyframes which are connected to each other to form a pose graph.
For this work, we extend UPSLAM to additionally integrate semantic labels extracted from imagery to form semantic panoramas.  Notably, UPSLAM does not use the semantics for scan-matching at all, it simply uses the estimated rigid body transforms to integrate the semantic information.  Therefore, because we do not require semantic data on every scan, we can run our inference at a rate lower than the LiDAR without removing data for ICP to improve map quality.  Sample depth, normal, and semantic panoramas are shown in Fig.~\ref{fig:panos}.

In addition to using UPSLAM egomotion estimates for the particle filter motion model, we compute the covariance and mean of the posterior particle filter estimate at each update. Once the covariance drops below a threshold, $\Sigma_t$, we use the estimated position with respect to the overhead map as a prior factor in the pose graph to anchor the corresponding panorama to the global map.  The final resulting pose graph is shown in Fig.~\ref{fig:block_diag}.
In this way, we allow the graph optimization to also georeference the global map based on the global localization.  Our experiments show that adding these semantic edges removes drift, effectively creating semantic loop closures to the global map at every keyframe.  This enables the mapper to handle much larger trajectories without loops while remaining globally consistent.

\section{Evaluation Datasets}

We first evaluate our system on SemanticKITTI \cite{behley_iccv_2019}, a manually point-wise labelled dataset of LiDAR scans from the KITTI odometry benchmark \cite{geiger_cvpr_2012}.  In this work, we test on runs 00, 02, and 09 from the odometry dataset.

In addition to SemanticKITTI, we gather our own datasets in Philadelphia and Morgantown, PA in order to test on a greater diversity of environments. These datasets were gathered using an Ouster OS-1-64 LiDAR along with 4 PointGrey Flea3 RGB cameras.  The cameras are hardware synchronized to the Ouster and calibrated using our method presented in \cite{shivakumar_itsc_2019}.  Table~\ref{tab:dataset_summary} gives an overview of the datasets we use for evaluation.  We also use an additional dataset \texttt{ucity'}, which follows the same trajectory as \texttt{ucity} but was taken several months prior.  We additionally start \texttt{kitti2} 50 seconds in and \texttt{kitti9} 10 seconds in order to avoid regions where UPSLAM fails at the beginning of the datasets.

\begin{table}[]
    \centering
    \vspace{2mm}
    \begin{tabular}{c||c|c|c|c|c}
        Dataset & kitti0 & kitti2 & kitti9 & morg & ucity \\
        \hline 
        Location & Karlsruhe & Kar. & Kar. & Morg. & Phil. \\
        Duration (sec) & 366 & 335 & 89 & 749 & 967 \\
        Length (m) & 2999 & 3581 & 959 & $\sim$6500 & $\sim$8500 \\
        Scale (px/m) & 3.0671 & 1.627 & 2.905 & 1.31 & 1.097 \\
        PC labelling & PC & PC & PC & RGB & RGB \\
        Sat. labelling & Man. & Auto & Auto & Man. & Man.
    \end{tabular}
    \caption{Experimental datasets used.  PC labelling indicates if inference was performed on LiDAR (PC) or RGB images (RGB).  Sat. labelling designates whether the top-down image was labelled by hand or by a network.}
    \label{tab:dataset_summary}
    \vspace{-3mm}
\end{table}

\section{Results}

\subsection{Inference Performance}
We first evaluate the performance of our inference networks, the results of which are shown in Table~\ref{tab:inf_iou}.  For the LiDAR network, the test set was KITTI 00, 02, and 09.  For the aerial network, the model was trained on 3/4 of each training image; the final 1/4 was used for evaluation.  We note that this does not well-test generalization to other cities, but robust multi-city aerial segmentation is not the focus of this work.  However, we emphasize that our localizer is able to perform well with significant realistic noise in the segmentation data, both in the LiDAR scans as well as the aerial imagery.

\begin{table}[]
    \centering
    \begin{tabular}{c||c|c|c|c|c|c|c}
         & Avg & Vehicle & Veg & Road & Ter. & Bldg & Other \\
         \hline
         LiDAR & 0.83 & 0.90 & 0.74 & 0.89 & 0.74 & 0.76 & 0.89 \\
         Aerial & 0.60 & N/A & 0.77 & 0.54 & 0.48 & 0.62 & N/A
    \end{tabular}
    \caption{Per-class Intersection over Union for aerial and LiDAR segmentation networks.}
    \label{tab:inf_iou}
    \vspace{-5mm}
\end{table}

\begin{figure}
    \centering
    \vspace{2mm}
    \includegraphics[width=0.95\linewidth]{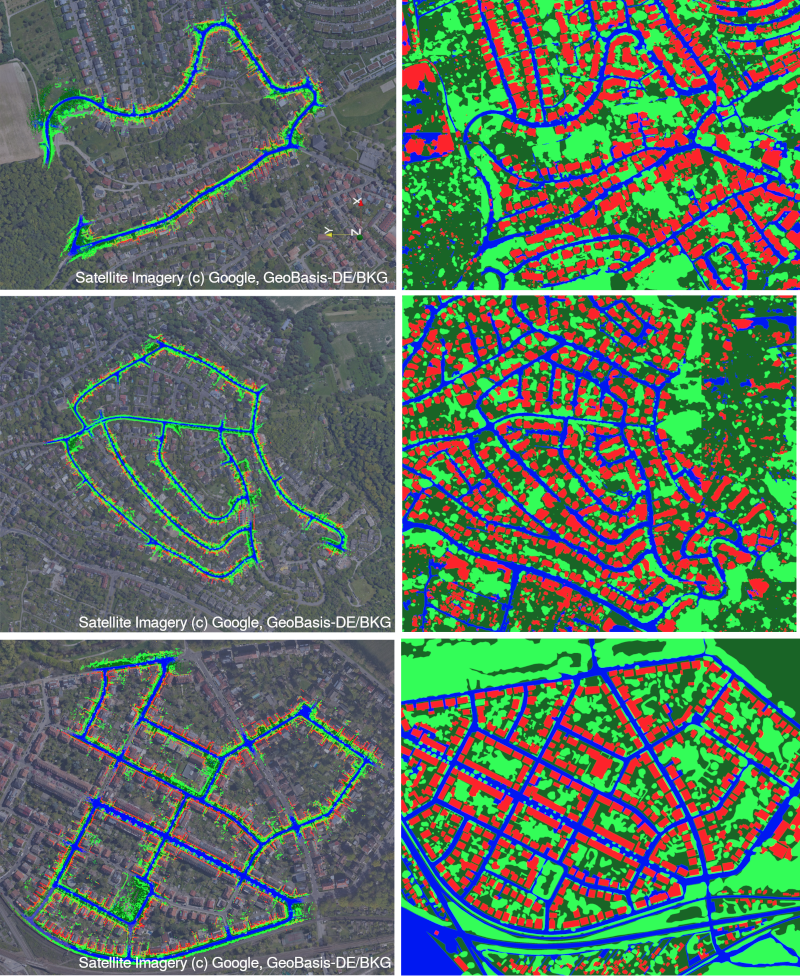}
    \caption{Maps overlain on satellite images for KITTI datasets 9, 2, and 0 (top to bottom).  The top-down segmentations are shown as well.  Datasets 9 and 2 are automatically labelled by our satellite segmentation network.}
    \label{fig:labelled_maps}
    \vspace{-3mm}
\end{figure}

\subsection{Accuracy}
We test on KITTI 00, 02, and 09 by running each trajectory 5 times, both with active scale estimation as well as using the fixed ground-truth scale. The resulting maps from one such run are shown in Fig.~\ref{fig:labelled_maps}.

\begin{figure}
    \vspace{2mm}
    \centering
    \includegraphics[width=0.78\linewidth]{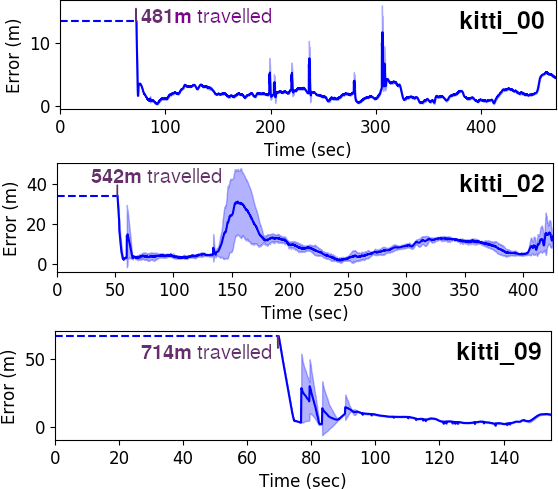}
    \caption{Errors for 5 different runs of different KITTI datasets with standard deviation shown.  The time and distance travelled before convergence are shown.}
    \label{fig:kitti_err}
    \vspace{-5mm}
\end{figure}

Fig.~\ref{fig:kitti_err} shows the position error in pixels throughout the entire dataset after convergence was automatically detected.  For these runs, the ground truth scale was provided to the system.  All parameters were fixed for all runs except for \texttt{kitti9}, for which we increased the angular process noise by a factor of 10 as UPSLAM's estimates were noisier.  The average errors were 2.0, 9.1, and 7.2 meters respectively. For the later datasets, the error was concentrated in several local regions which were typically regions where UPSLAM struggled (generally in tunnel-like regions of trees with less geometric structure) so the motion priors were not as accurate.  We compare these results to \cite{kim_icra_2017} and \cite{yan_ecmr_2019} in Table~\ref{tab:method_comparison}. By incorporating much richer semantic data, we are able to outperform \cite{yan_ecmr_2019} by a factor of 10 on \texttt{kitti0}.  Kim et al.~\cite{kim_icra_2017} obtain localization accuracies comparable to ours, though using a single RGB camera instead of LiDAR and requiring the precomputation of deep embeddings on a grid over the satellite image.

\begin{table}[]
    \centering
    \begin{tabular}{c||c|c|c}
        Method & \texttt{kitti0} & \texttt{kitti2} & \texttt{kitti9} \\
        \hline
        Ours (scale fixed) & 2.0m/56.4sec & 9.1m/71.5sec & 7.2m/$\sim$75sec \\
        Yan et al.~\cite{yan_ecmr_2019} & $\sim$20m/$\sim$60sec & N/A & $\sim$25m/$\sim$17sec \\
        Kim et al.~\cite{kim_icra_2017} & $\sim$4.6m/$\sim$62sec & N/A & $\sim$7.7m/$\sim$55.6sec \\
    \end{tabular}
    \caption{Mean accuracy and convergence time (shown as acc/conv) for the KITTI datasets with various methods.  N/A indicates data not given in the papers. For \cite{kim_icra_2017}, accuracy is given for the end of the dataset only.}
    \label{tab:method_comparison}
    \vspace{-5mm}
\end{table}

These experiments on KITTI were performed entirely in real-time, including segmentation, localization, and mapping, running on an AMD Ryzen 9 3900X CPU and NVidia GeForce RTX 2080 GPU. 
This level of computing can readily be installed on a vehicle or a moderately sized ground robot.
We also expect that with further optimizations of our method, such as moving the particle weighting function to the GPU, we could run in real-time on platforms with more constrained computational resources.

\begin{figure}
    \vspace{2mm}
    \centering
    \includegraphics[width=0.8\linewidth]{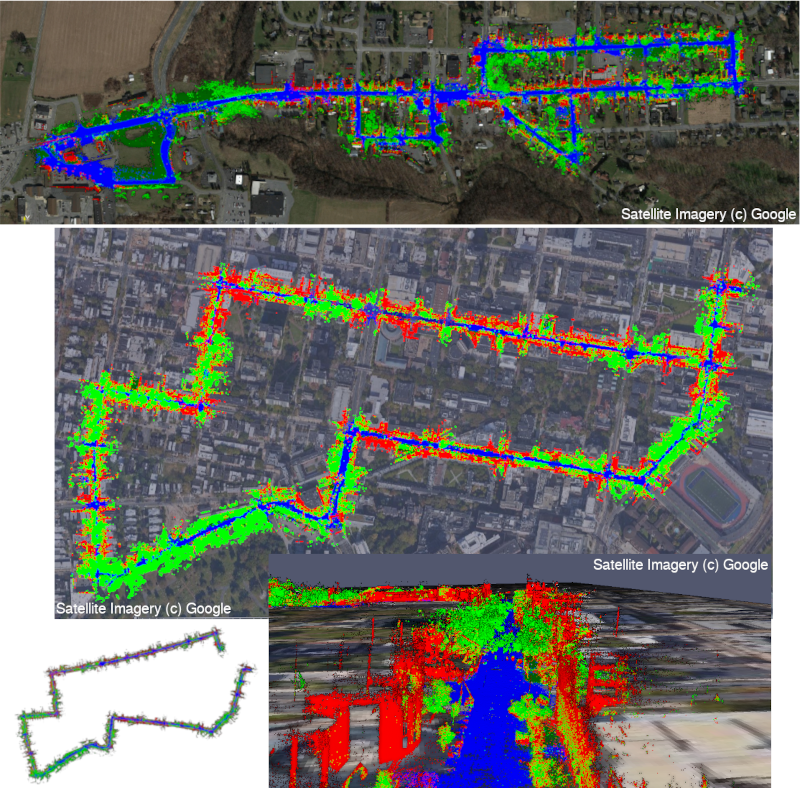}
    \caption{Top view and side view of \texttt{morg} and \texttt{ucity'} semantic maps overlaid onto the satellite imagery.  The lower left shows \texttt{ucity} built without using semantic localization.}
    \label{fig:ucity_map_and_eye}
    \vspace{-5mm}
\end{figure}


In addition, we qualitatively inspect the results for \texttt{ucity'} and \texttt{morg} in Fig.~\ref{fig:ucity_map_and_eye}.  We note that by incorporating global edges into the mapper pose graph we are able to not only globally localize the robot in the top-down image, but also improve global map consistency.  Over the course of long loops, UPSLAM by itself is unable to find loop closures due to drift, resulting in globally inconsistent maps.  Incorporating our global edges from the semantic localizer mitigates this drift, allowing UPSLAM to find the loop closures successfully.

\subsection{Convergence}

\begin{figure}
    \vspace{2mm}
    \centering
    \includegraphics[width=0.62\linewidth]{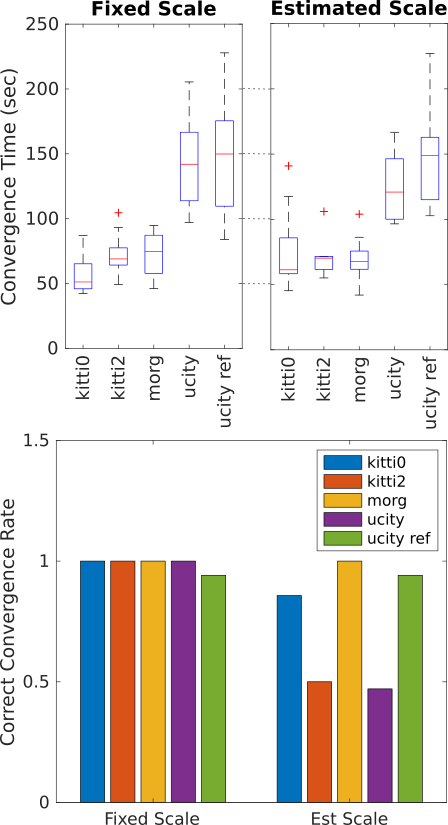}
    \caption{Convergence robustness and time to convergence for different datasets, both with and without active scale estimation.}
    \label{fig:conv_analysis}
    \vspace{-5mm}
\end{figure}

We perform a convergence analysis on the UCity, Morgantown, and KITTI 0 and 2 datasets by starting the run at even intervals throughout the dataset.  We do not analyze \texttt{kitti9} as it is relatively short.  We define the correct convergence rate to be the fraction of detected convergences where the average error for the first 20 seconds after convergence is less than 10 meters.  Fig.~\ref{fig:conv_analysis} shows the frequency of correct convergences to overall detected convergences.  We ignore runs where the particle filter did not converge before the end of the run, which occurred for starting times where the time remaining in the dataset was less than the convergence time.  Identical parameters were used for all runs with the exception of doubling the regularization term $\gamma$ for the fixed-scale case.

\begin{figure}
    \centering
    \includegraphics[width=0.65\linewidth]{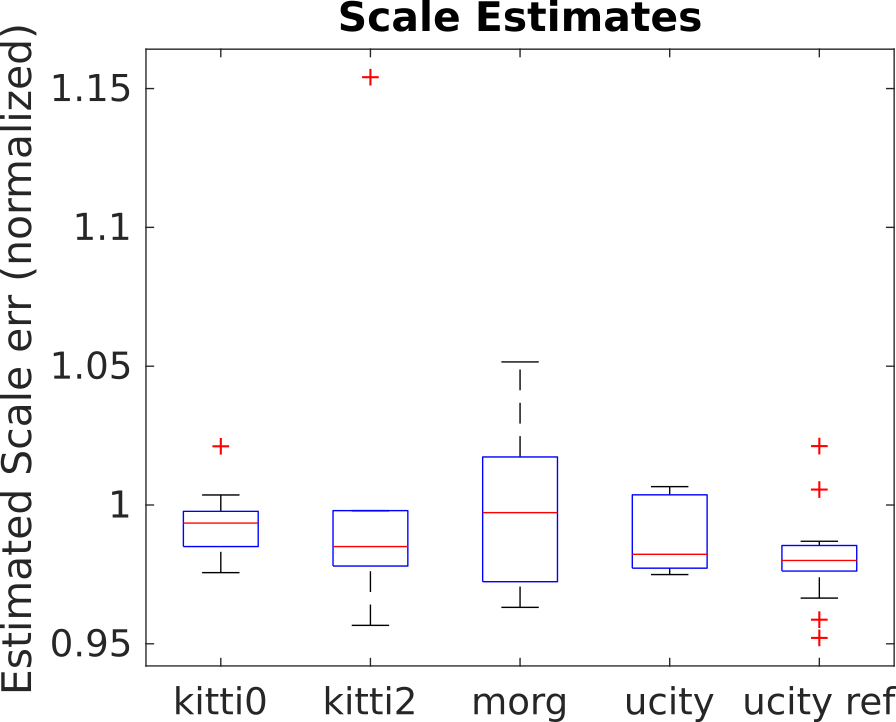}
    \caption{Estimated scale distribution from runs which converged correctly in terms of $s_{est}/s_{true}$.}
    \label{fig:scale_est}
    \vspace{-5mm}
\end{figure}

\begin{figure}
    \vspace{2mm}
    \centering
    \includegraphics[width=0.9\linewidth]{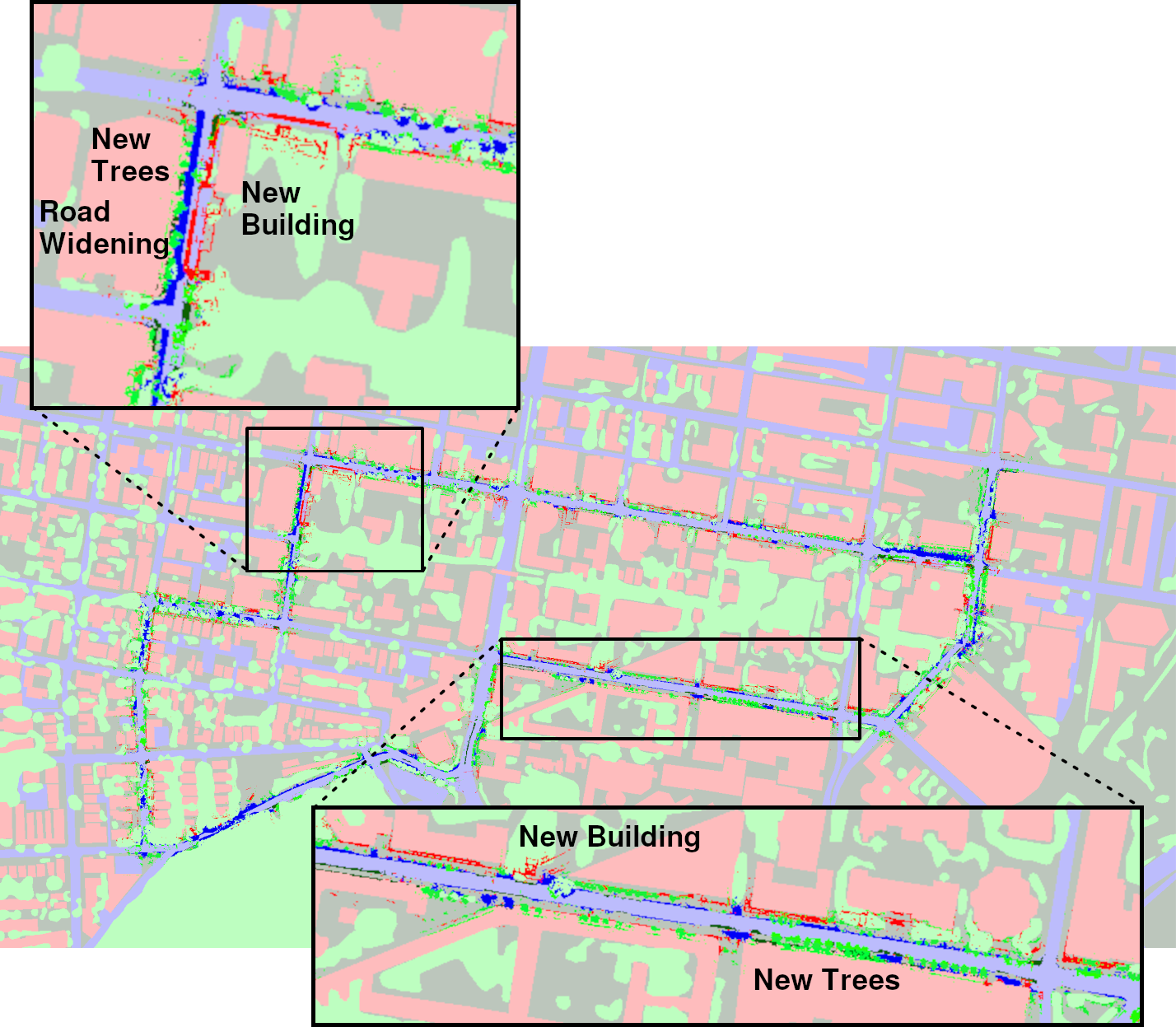}
    \caption{The refined top-down segmentation of \texttt{ucity'}, with unchanged portions shown faded. Note the addition of under-construction buildings not in the satellite image.}
    \label{fig:ucity_refined}
    \vspace{-3mm}
\end{figure}

From these results we conclude that convergence is more difficult when estimating scale, which holds with our intuition.  We use the same number of particles for both cases, making sample impoverishment more likely.  \texttt{ucity} is notably significantly harder for our system, which we attribute to the environment being much closer to a uniform grid with less distinctive structure to aid localization. In order to improve localization, we run our localizer on a dataset of the same trajectory (\texttt{ucity'}) taken several months prior and use the generated map to refine the original satellite segmentation.  The result of this process is shown in Fig.~\ref{fig:ucity_refined}.  Using this new refined map \texttt{ucity ref} we are able to significantly improve the correct convergence rate, indicating that the prior run, despite being several months earlier, was able to add useful context to the map.  Kim et al.~\cite{kim_icra_2017} and Yan et al.~\cite{yan_ecmr_2019} obtain comparable times in Table~\ref{tab:method_comparison}, but are unable to estimate scale. For \texttt{kitti9} we only start at the beginning due to the short dataset.  Note that these works compute convergence times from the start of the dataset only, whereas we average times from starting at various points.

Finally, analysis of the estimated scale from our system in Fig.~\ref{fig:scale_est} shows that our method can estimate scale accurately within approximately a factor of 0.05 of the true value in most cases.

\subsection{Ablation Study}

\begin{figure}
    \centering
    \includegraphics[width=0.95\linewidth]{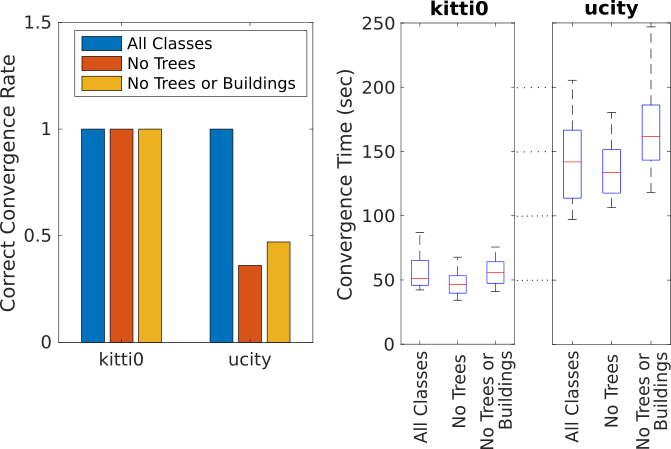}
    \caption{Convergence rates and times for different datasets when ignoring trees and/or buildings in particle weighting.}
    \label{fig:conv_analysis_ablations}
    \vspace{-5mm}
\end{figure}

Noting that map refinement was so helpful for improving convergence for \texttt{ucity}, we investigated how much our localizer was relying simply on road shape and trajectory and how much it was using additional cues from buildings and trees.  We ablate the cost from trees and buildings and report these findings in Fig.~\ref{fig:conv_analysis_ablations}.  For all runs we assumed a fixed known scale.  We conclude that for environments and trajectories with complex road structure, simply utilizing the trajectory and matching it to the road pattern is sufficient.  However, for more Manhattan-like structure, local details such as buildings and trees become more relevant.  Interestingly, for \texttt{ucity}, removing just trees was slightly worse than both buildings and trees.  We hypothesize that this is because in a dense urban environment, there are almost always buildings on both sides, so they add very little structure beyond roads, and in the case of noise can be detrimental.  For \texttt{ucity}, however, the trees evidently provided strong helpful cues. 

\section{Conclusion}
We have presented a system for real-time semantic localization and mapping in satellite or top-down drone imagery. In addition, we have shown quantitative and qualitative results on multiple datasets gathered with different sensors in a variety of environments. In all of these experiments, our system performs robustly and accurately. In the future, we plan to deploy our system on a heterogeneous team of robots to enable decentralized mapping and higher-level autonomy.

\bibliographystyle{IEEEtran}
\bibliography{bib/bib}

\end{document}